\documentclass[10pt,twocolumn,letterpaper]{article}

\usepackage[pagenumbers]{cvpr} 

\usepackage{ascii}
\usepackage[utf8]{inputenc} 
\usepackage[T1]{fontenc}    
\usepackage{url}            
\usepackage{booktabs}       
\usepackage{amsfonts}       
\usepackage{nicefrac}       
\usepackage{microtype}      
\usepackage[dvipsnames,svgnames,x11names]{xcolor} 
\usepackage{float}

\usepackage{enumerate}
\usepackage{ntheorem}
\usepackage{makecell}

\usepackage{colortbl}
\usepackage{enumitem}
\usepackage{amsmath}
\usepackage{amssymb}
 \usepackage[pdftex]{graphicx} 
\usepackage{pifont}
\usepackage{wasysym}
\usepackage{mathrsfs}
\usepackage{multirow}
\usepackage{textcomp}
\usepackage{footnote}
\usepackage{soul}
\usepackage{threeparttable}
\usepackage[ruled,vlined]{algorithm2e}
\usepackage{color}
\definecolor{mygray}{gray}{.9}
\definecolor{light-gray}{gray}{0.5}
\definecolor{pretty-blue}{RGB}{0, 113, 188}
\definecolor{linecolor}{RGB}{255,255,255}
\definecolor{linecolor1}{RGB}{255,255,255}
\definecolor{kaiming-green}{RGB}{57,181,74} 
\definecolor{icmlblue}{rgb}{0,0.08,0.45} 

\def\etal{{\it{et al.}}}

\usepackage{xspace}

\definecolor{myblue}{rgb}{.39,.58,.93}

\usepackage{pifont}

\definecolor{prompt_blue}{HTML}{1f78b4}
\definecolor{prompt_red}{HTML}{d45c43}

\usepackage[colorlinks=True,
            linkcolor=red,
            anchorcolor=blue,  
            pagebackref,
            breaklinks=True,
            ]{hyperref}

\usepackage[capitalize]{cleveref}
\crefname{section}{Sec.}{Secs.}
\Crefname{section}{Section}{Sections}
\Crefname{table}{Table}{Tables}
\crefname{table}{Tab.}{Tabs.}

\def\cvprPaperID{****} 
\def\confName{CVPR}
\def\confYear{2023}

\begin{document}

\title{Multi-Glimpse Network: A Robust and Efficient Classification Architecture\\ based on Recurrent Downsampled Attention}

\author{Sia Huat Tan\\
Tsinghua University\\
{\tt\small csf19@mails.tsinghua.edu.cn}
\and
Runpei Dong\\
Xi'an Jiaotong University\\
{\tt\small runpei.dong@gmail.com}
\and
Kaisheng Ma\\
Tsinghua University\\
{\tt\small kaisheng@mail.tsinghua.edu.cn}
}
\maketitle

\begin{abstract}
    Most feedforward convolutional neural networks spend roughly the same effort for each pixel.
    Yet human visual recognition is an interaction between eye movements and spatial attention, which we will have several glimpses of an object in different regions.
    Inspired by this observation, we propose an end-to-end trainable \textbf{M}ulti-\textbf{G}limpse \textbf{N}etwork (\textbf{MGNet}) which aims to tackle the challenges of high computation and the lack of robustness based on recurrent downsampled attention mechanism.
    Specifically, MGNet sequentially selects task-relevant regions of an image to focus on and then adaptively combines all collected information for the final prediction.
    MGNet expresses higher resistance against adversarial attacks and common corruptions with less computation. 
    Also, MGNet is inherently more interpretable as it explicitly informs us where it focuses during each iteration.
    Our experiments on ImageNet100 demonstrate the potential of recurrent downsampled attention mechanisms to improve a single feedforward manner.
    For example, MGNet improves 4.76\% accuracy on average in common corruptions with only 36.9\% computational cost.
    Moreover, while the baseline incurs an accuracy drop to 7.6\%, MGNet manages to maintain 44.2\% accuracy in the same PGD attack strength with ResNet-50 backbone.
    Our code is available at \texttt{\url{https://github.com/siahuat0727/MGNet}}.
\end{abstract}

\section{Introduction}
\label{intro}

Convolutional Neural Networks (CNNs) have achieved promising performance on many visual tasks, such as object detection~\cite{girshick2014rich, ren2015faster, redmon2016you}, image segmentation~\cite{long2015fully, chen2017deeplab} and image captioning~\cite{DBLP:journals/corr/abs-1805-09019, DBLP:conf/acl/DevlinCFGDHZM15, DBLP:journals/corr/abs-1810-04020}.
Especially in image classification~\cite{krizhevsky2017imagenet, huang2017densely, DBLP:conf/bmvc/ZagoruykoK16}, CNNs can even surpass human performance~\cite{DBLP:conf/iccv/HeZRS15, he2016deep}.

However, CNNs are facing various challenges: 1) CNNs are computationally expensive and memory intensive.
This increases the difficulty for CNNs to be widely deployed in scenarios like edge-computing;
2) CNNs are vulnerable to adversarial example~\cite{nguyen2015deep, goodfellow2014explaining, szegedy2013intriguing}, which is usually an image formed by making a subtle perturbation that leads a trained model to produce an incorrect prediction.
This raises major concerns about deploying neural networks in high-security-demanding systems;
3) CNNs will be confused by many forms of common corruptions~\cite{hendrycks2019benchmarking}, such as bad weather, noise, and blur.
The lack of robustness is hindering some processes like autonomous vehicle development~\cite{DBLP:journals/corr/abs-1912-10773}.

\begin{figure}[t!]
\begin{center}
\includegraphics[width=\linewidth]{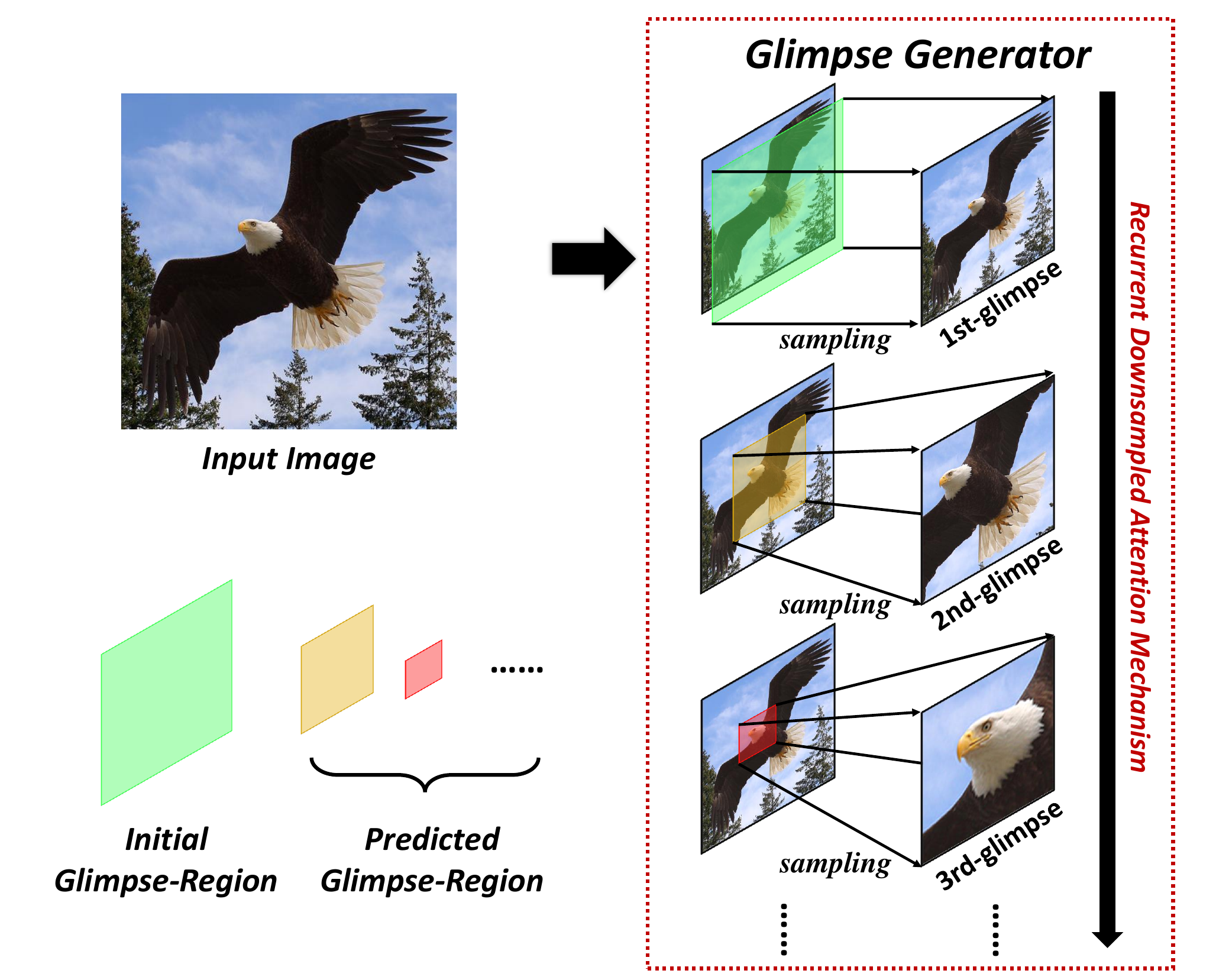}
\end{center}
   \caption{
   Illustration of the recurrent downsampled attention mechanism. From top to down, the Glimpse Generator sequentially generates glimpses by sampling from the given glimpse-regions in a recurrent manner.
   }
\label{fig:sample}
\end{figure}

Inspiration from the human visual system is a potential hint to solve both the expensive computation and robustness problems.
A particularly striking difference between the human visual system and current feedforward convolutional neural networks (FF-Nets) is that the FF-Nets spend enormous and roughly the same amount of computational energy on every single pixel, no matter whether it is essential to the task.
Additionally, most FF-Nets process the entire scene just once.
The human visual system, by contrast, is not merely feedforward but has various feedback and recurrent connections in the visual cortex~\cite{vision20}.
In addition, human beings don't treat an image as a static scene.
Instead, cognitive processing is an interaction between attention and eye movements~\cite{Saccadic}.
Specifically, the fovea in the human's eye samples distinct regions of the scene at varying spatial resolutions~\cite{Primate}.
The series of fixation on different locations and resolutions are then collected and integrated to build up an internal representation of the scene~\cite{Dynamic}.

Inspired by the sequential and variable resolution sampling mechanisms in the human visual system, we propose Recurrent Downsampled Attention (RDA) mechanism and present a novel Multi-Glimpse Network (MGNet) to explore the benefits of deploying RDA in CNNs.
Instead of sweeping the entire scene at once, our model sequentially selects to focus on some task-relevant regions (illustrated in Figure~\ref{fig:sample}).
During each iteration, our model will first apply variable resolution sampling to various size regions of the original image to produce a much lower dimensionality fixation, which we will refer to as glimpse~\cite{Glimpse}.
Every glimpse will be integrated over time to build up a global internal representation.
Since our model only mainly computes on these low dimensionality glimpses, the model can save computational costs.
Unlike other model acceleration methods, such as network pruning~\cite{han2015deep}, knowledge distillation~\cite{hinton2015distilling}, quantization~\cite{courbariaux2016binarized,han2015deep}, and model compacting~\cite{DBLP:conf/cvpr/SandlerHZZC18}, we break the current paradigm that sweeps the image just once and predicts.
By sequentially processing multiple glimpses, we further show that our model is fundamentally more robust against adversarial attacks and common corruption.

Our main contributions can be summarized as follows:
\begin{itemize}
\item We propose Multi-Glimpse Network, which is end-to-end trainable in one stage while not requiring any supervised spatial guidance or hand-crafted pre-training method.
\item With the same amount of computational cost, we demonstrate that MGNet outperforms FF-Nets with various backbones. Additionally, as the network is shared over iterations, it can decide to early-exit on-the-fly without adding any overhead.
\item We show that MGNet is intrinsically more robust against adversarial attacks and common corruptions.
For example, accuracy is improved by 4.76\% in common corruptions with 36.9\% computational requirement on average.
\end{itemize}

\section{Related Work}

\textbf{Robustness.}
Szegedy \textit{et al.}~\cite{szegedy2013intriguing} first show that a carefully perturbed image can fool a trained model entirely in high confidence. 
Goodfellow \textit{et al.}~\cite{goodfellow2014explaining} propose FGSM to generate adversarial examples.
Madry \textit{et al.}~\cite{madry2017towards} study the adversarial robustness of neural networks and propose a robust minimax optimization called PGD adversarial training. The research direction in studying  adversarial attack and defense method is in the progress~\cite{DBLP:journals/ijautcomp/XuMLDLTJ20, DBLP:journals/corr/abs-2002-07405, DBLP:conf/iccv/Mustafa0HGS019}.
Besides, Hendrycks and Dietterich~\cite{hendrycks2019benchmarking} consider common real-world corruptions and propose a benchmark to measure general robustness.
Recently, various data augmentation techniques~\cite{geirhos2018imagenet, hendrycks2019augmix, hendrycks2020many} are introduced to improve the general robustness.

\noindent\textbf{Computational Efficiency.}
Many research work have been proposed to reduce the computational cost of deep neural networks.
As there are considerable redundant parameters in neural networks, some focus on pruning the non-essential connections to reduce computational cost~\cite{han2015deep, DBLP:conf/bmvc/SrinivasB15, DBLP:conf/iclr/UllrichMW17}.
Another approach is quantization, which focuses on compressing the bit-width of weights for floating-point operations and memory usage reduction~\cite{DBLP:conf/eccv/RastegariORF16, DBLP:conf/iclr/ChoiEL17}.
Hinton \textit{et al.}~\cite{hinton2015distilling} propose knowledge distillation where the student learns to mimic the teacher's prediction results.
This technique has been widely used to transfer
the knowledge from larger models into compact models~\cite{DBLP:journals/corr/RomeroBKCGB14, DBLP:conf/aaai/LuoZLWT16}.
Recent works further reduce computation by designing efficient network architectures~\cite{DBLP:journals/corr/HowardZCKWWAA17, DBLP:journals/pami/HuSASW20, DBLP:conf/eccv/MaZZS18}.

\noindent\textbf{Recurrent Attention Model.} 
Recurrent attention mechanism has been explored in many fields, such as reinforcement learning~\cite{niv2015reinforcement, DBLP:conf/cvpr/HaqueAF16}, machine translation~\cite{DBLP:conf/icml/GehringAGYD17, DBLP:journals/corr/BahdanauCB14, NMT}, image classification~\cite{RAM, zoran2019robust, DBLP:journals/corr/abs-2103-03206} and generative models~\cite{DBLP:journals/corr/abs-1805-08318, DBLP:conf/nips/LiaoLSWHDUZ19, DBLP:conf/nips/EslamiHWTSKH16}.
In the vision task, Mnih \textit{et al.}~\cite{RAM} first propose a recurrent visual attention model to control the amount of computation on the augmented MNIST dataset~\cite{lecun-mnisthandwrittendigit-2010}. While the model is not differentiable, it is trained using reinforcement learning.
Gregor \textit{et al.}~\cite{gregor2015draw} propose differentiable attention mechanisms to generate images sequentially.
Jaderberg \textit{et al}.~\cite{jaderberg2015spatial} show that meaningful object parts can be discovered automatically with only image labels.
Fu \etal ~\cite{Zheng_2017_ICCV} propose a recurrent attention model
to learn region-based feature representation at multiple scales in fine-grained image classification.
Zoran \textit{et al.}~\cite{zoran2019robust} show that an adversarially trained sequential attention network is significantly more robust than a feedforward model.

Since each recurrent attention-related work has a different focus, most of them are designed experimentally using multiple model capacity or computational cost or both.
In this work, with the proposed RDA and MGNet, we aim to answer a question:
\textit{given the same model capacity and computational cost, is it beneficial to introduce recurrent mechanism in CNNs?}
Our experiments further show that MGNet is intrinsically more robust against adversarial attacks, and a low-dimensionality glimpse is crucial to improve general robustness.

\section{Approach}

\begin{figure*}[t]
\begin{center}
   \vspace{10pt}

\includegraphics[width=1.0\linewidth]{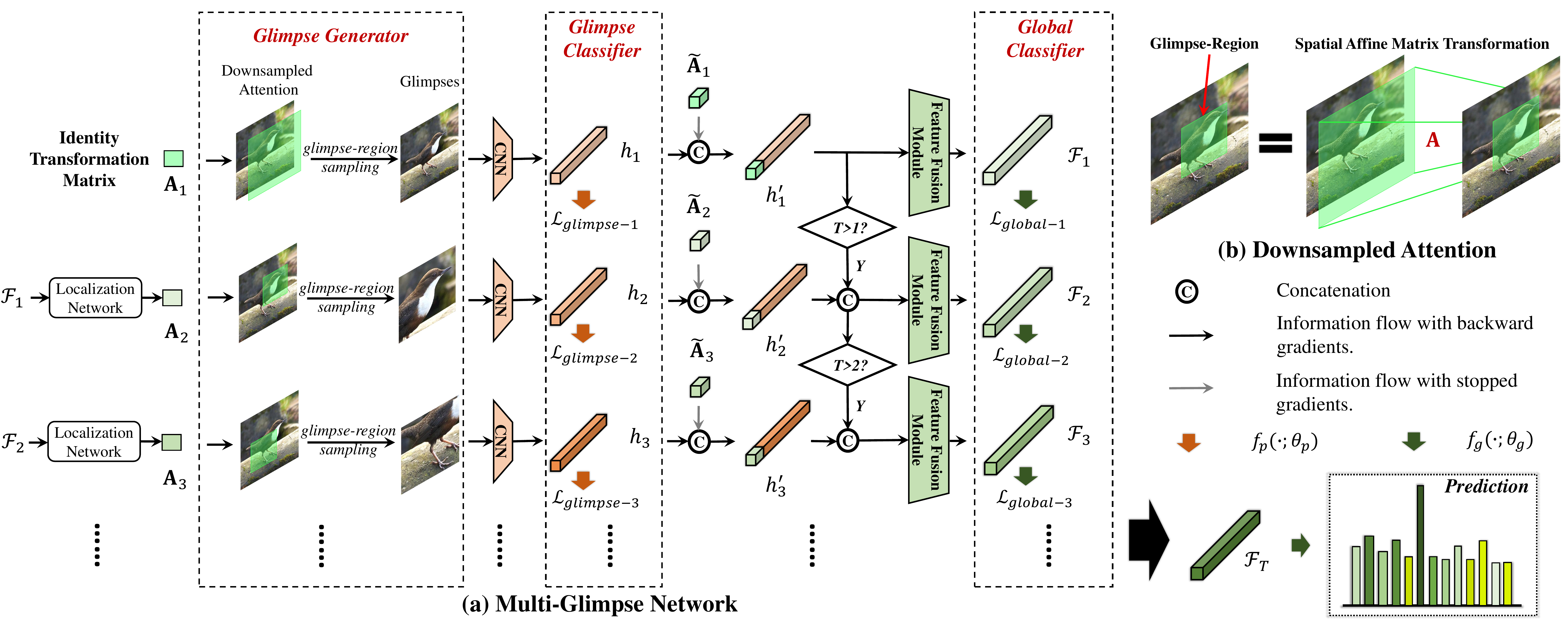}
\end{center}
   \caption{
   Details of our method: (a) The framework of MGNet.
   Glimpses are generated by sequentially sampling the image from the glimpse-region.
   The Glimpse Classifier guides to make every glimpse count and will be dropped after training.
   Multi glimpse features are integrated by Feature Fusion Module into a global feature.
   The global feature will be decoded to predict the label and the next glimpse-region.
   Note that we share all the parameters during the iterations.
   (b) Illustration of the downsampled attention.
   \textit{Best viewed in color.}
   }
  \vspace{-10pt}

\label{fig:structure}
\end{figure*}

In this section, we present an overview of our proposed MGNet, as illustrated in Figure ~\ref{fig:structure}.
Instead of blindly carrying out a large amount of computation for every single pixel of an image, our model will sequentially generate $T$ glimpses to be processed and fuse all the glimpses for the final prediction.

Given an image $\boldsymbol{x} \in \mathbb{R}^{H \times W}$, $H$ and $W$ respectively denote the height and width of the image. For the $t$-th iteration, the Glimpse Generator $g$ will apply an affine transformation to the input image and perform sampling to produce a glimpse $\boldsymbol{x}^g_t = g(\boldsymbol{x}, {\mathbf{A}}_t; M)$, where $\boldsymbol{x}^g_t \in \mathbb{R}^{\frac{H}{M} \times \frac{W}{M}}$, $M$ is a downsampling factor and ${\mathbf{A}}_t$ is the $t$-th affine transformation matrix.
The downsampling factor $M$ is fixed and greater than $1$ to reduce the amount of computation.
${\mathbf{A}}_t$ is generated by the Localization Network, except for the initial matrix ${\mathbf{A}}_1$, which we set as an identity transformation matrix. Therefore, the first glimpse will be a low-resolution version of the original image. We will introduce the Glimpse Generator in Section~\ref{gg}.

The glimpse $\boldsymbol{x}^g_t$ is first encoded by a CNN backbone  (including global average pooling) to produce a glimpse feature $h_t$.
Each glimpse feature will be decoded by a glimpse classifier $f_p(\cdot;\theta_p)$ into class logits to make every glimpse count.
The affine transformation matrix ${\mathbf{A}}_t$ will be flattened as $\Tilde{\mathbf{A}}_t$ and appended to the glimpse feature $h_t$.
We stop the gradient on $\Tilde{\mathbf{A}}_t$ as it is a positional encoding that can help the model understands where the feature comes from.
Then all glimpse features will be integrated by a Feature Fusion Module to produce global internal representation $\mathcal{F}_t$ of the image during the $t$-th iteration. This module will be introduced in Section~\ref{ff}.

With a fully-connected layer $f_g(\cdot;\theta_g)$ as the global classifier, we decode $\mathcal{F}_t$ into class logits iteratively to produce $T$ classification results.
Note that the decoded result of $\mathcal{F}_T$ will be the final prediction.
$\mathcal{F}_t$ will also be fed into the Localization Network to generate the next glimpse-region (if needed), and more details can be found in Section~\ref{ln}.

\subsection{Glimpse Generator} \label{gg}

This module aims to generate low-dimensionality glimpses.
The non-differentiability of cropping and resizing makes it difficult to learn where to look, which can be addressed with reinforcement methods such as policy gradient~\cite{RAM}.
We will briefly introduce a differentiable affine transformation operation proposed by Jaderberg \textit{et al}.~\cite{jaderberg2015spatial}, making it possible to be trained end-to-end with SGD.

We first generate a 2D flow field (we call it glimpse-region) by applying a parameterized sampling grid with an affine transformation matrix $\mathbf{A}$.
Since we only consider cropping, translation, and isotropic scaling transformations, ${\mathbf{A}}$ is more constrained and requires only 3 parameters,

\begin{equation} \label{eq1}
{\mathbf{A}}  = 
\begin{bmatrix}
a^s & 0 & a^x \\
0 & a^s & a^y
\end{bmatrix},
\end{equation}

\noindent where $a^s$, $a^x$, and $a^y$ are the output of the Localization Network (details in Section~\ref{ln}).

To generate a glimpse $\boldsymbol{x}^g$, we first perform a pointwise transformation
\begin{equation}
    \begin{pmatrix}
        {x_{t_x}} \\
        {x_{t_y}}
    \end{pmatrix}
    = {\mathbf{A}}  
    \begin{pmatrix}
        {x_{t_x}^g} \\
        {x_{t_y}^g} \\
        1
    \end{pmatrix},
\end{equation}

\noindent where (${x_{t_x}}$, ${x_{t_y}}$) are the coordinates of the regular grid in the input image $\boldsymbol{x}$, and (${x_{t_x}^g}$, ${x_{t_y}^g}$) are the coordinates that define the sample points.
Then we apply a bilinear sampling to generate a glimpse $\boldsymbol{x}^g \in \mathbb{R}^{\frac{H}{M} \times \frac{W}{M}}$.
Especially, for the first glimpse, we let $a^s$ equal to 1 and $a^x$, $a^y$ equal to 0, which denote an identity transformation.
Since the downsampling factor $M$ is greater than 1, the first glimpse represents a low-resolution version of the input image.
The differentiability of this affine transformation allows our model to learn the task-relevant regions with backpropagation.

\subsection{Feature Fusion Module} \label{ff}

It is crucial to integrate the information of every glimpse to make the final prediction.
In this section, we introduce our Feature Fusion Module, using attention mechanism~\cite{vaswani2017attention} with a single attention head, to integrate all the glimpse features $h_1, h_2, \cdots , h_t$ into a global internal feature $\mathcal{F}_t$.
Specifically, for the $t$-th iteration, 
\begin{equation}
    \begin{split}
        \mathbf{H}_t &= \mathrm{concatenate}\left([{h_1}^\prime, {h_2}^\prime, \cdots, {h_t}^\prime]\right), \\
        {\mathcal{E}_t} &= \mathrm{softmax}\left(\frac{(\mathbf{H}_t \mathbf{W}^q)(\mathbf{H}_t \mathbf{W}^k)^\mathsf{T}}{\sqrt{d}}\right)
        (\mathbf{H}_t \mathbf{W}^v) \mathbf{W}^o, \\
        \mathcal{F}_t &= \mathrm{ReLU}\big(\mathrm{LayerNorm}({\mathcal{E}_t})[t]\big),
    \end{split}
\end{equation}
where ${h_t}^\prime \in \mathbb{R}^d$ is the glimpse feature $h_t$ concatenated with the positional encoding, $\mathbf{W}^q, \mathbf{W}^k,$
$\mathbf{W}^v, \mathbf{W}^o \in \mathbb{R}^{d \times d}$ are the learnable parameters, ${\mathcal{F}_t} \in \mathbb{R}^d$ is the global internal representation integrated during the $t$-th iterations, and the notation $\mathbf{X}[t]$ represents the $t$-th row of the matrix $\mathbf{X}$.
Note that for an experiment setting with $T$ iterations, $\mathcal{F}_T$ represents the final feature and will be decoded by the global classifier $f_g(\cdot;\theta_g)$ to predict the label.

\subsection{Localization Network} \label{ln}
 
We propose Localization Network to predict an affine transformation matrix ${\mathbf{A}}$ for the glimpse generation.
More intuitively, ${\mathbf{A}}$ can represent a target region of the input image, where the parameter $a^s$ is the ratio of the size of the glimpse-region to the input image, $a^x$ and $a^y$ denote the translation of the region origin.
In MGNet, we let $a^s \in [a^s_{\min}, a^s_{\max}]$ so that the glimpse-region size is adaptive, where $a^s_{\min} = 0.2$ and $a^s_{\max} = 0.5$.
Since we prevent the Glimpse Generator from sampling beyond the image range, the range of $a^x$ and $a^y$ should be within $[a^s-1, 1-a^s]$. In detail, given a $t$-th global internal representation $\mathcal{F}_{t}$, we produce the parameter of matrix $\mathbf{A}_{t+1}$ by
\begin{equation} \label{eq2}
\begin{split}
[{a^s_{t+1}}^\prime, {a^x_{t+1}}^\prime, {a^y_{t+1}}^\prime] &= \Phi\Big(\sigma \big( f_l(\mathcal{F}_{t}; \theta_l)\big); s \Big), \\
a^s_{t+1} &= {a^s_{t+1}}^\prime \cdot (a^s_{max} - a^s_{min}) + a^s_{min}, \\
[a^x_{t+1}, a^y_{t+1}] &= \big(2 \cdot[{a^x_{t+1}}^\prime, {a^y_{t+1}}^\prime] - 1\big) \cdot 
\big( 1 - a^s_{t+1} \big), 
\end{split}
\end{equation}
where $\sigma$ is sigmoid function,  $f_l(\cdot;\theta_l)$ is a fully-connected layer, $\Phi$ is a gradient re-scaling operation and $s$ is a gradient re-scaling factor.
The gradient re-scaling operation
\begin{equation}
    \Phi(\boldsymbol{x}; s) = \boldsymbol{x};\quad \nabla_{\boldsymbol{x}} \Phi(\boldsymbol{x}; s) = s
\end{equation}
is applied to tackle the gradient issue as we empirically find an exploding gradient problem in the Localization Network.
The value of $s$ is possibly around 0.01 to 0.02 in our setting.
We show the hyper-parameter tuning in Appendix \ref{hparam}.

\subsection{Joint Classifiers Learning}
Given dataset $D=\{\boldsymbol{x}^{(i)}, \mathbf{y}^{(i)}\}_{i=1}^{N}$ where $\boldsymbol{x}^{(i)}$ denotes the $i$-th input image and $\mathbf{y}^{(i)}$ is the corresponding label, 
MGNet jointly learns the glimpse feature together with the global internal feature in an end-to-end fashion. To  realistically 
demonstrate  the  model’s  potential,  we train our model on pure Cross-Entropy (CE) loss and hence the total loss $\mathcal{L}$ 
can be given as
\begin{equation}
    \mathcal{L} = \alpha \mathcal{L}_{glimpse} + (1-\alpha) \mathcal{L}_{global},
\end{equation}
\noindent where $\mathcal{L}_{glimpse}$ is the glimpse classifier loss, $\mathcal{L}_{global}$ is the global classifier loss, and $\alpha$ is
a hyper-parameter that balances the weighting between the losses.

As shown in Figure \ref{fig:structure}, the glimpse classifier can be regarded as an auxiliary loss and will be dropped after training, so neither extra memory nor computation power is required during inference. 
We show the effect of the glimpse classifier in Appendix \ref{hparam}.

\textbf{Global Classifier}.
During the $t$-th iteration, the global classifier takes $\mathcal{F}_t$ as input to make a global prediction, and it is trained by averaging all $t$-th prediction loss $\mathcal{L}_{global-t}$:

\begin{equation}
\begin{aligned}
    \mathcal{L}_{global} &= \frac{1}{T}\sum_{t=1}^T  \mathcal{L}_{global-t}, \text{where} \\
    \mathcal{L}_{global-t} &= \mathbb{E}_{x, y \sim D} \Big[\mathcal{H} \big(y, f_g(\mathcal{F}_t; \theta_g)\big)\Big],
\end{aligned}
\end{equation}

\noindent where $\mathcal{H}$ denotes the CE loss and $T$ is the number of glimpses.

\textbf{Glimpse Classifier}.
Similarly, the glimpse classifier takes $h_t$ as input and jointly learns by averaging all $t$-th glimpse loss $\mathcal{L}_{glimpse-t}$:

\begin{equation}
\begin{aligned}
    \mathcal{L}_{glimpse} = \frac{1}{T} \sum_{t=1}^T \mathcal{L}_{glimpse-t}, \text{where}\\
    {\mathcal{L}_{glimpse-t}} = \mathbb{E}_{x, y \sim D} \Big[\mathcal{H}\big(y, f_p(h_t; \theta_p)\big)\Big].
\end{aligned}
\end{equation}

\section{Experimental Results}

\subsection{ImageNet100}

\begin{figure*}[h]
\vspace{-15pt}
  \centering
  \includegraphics[width=\linewidth]{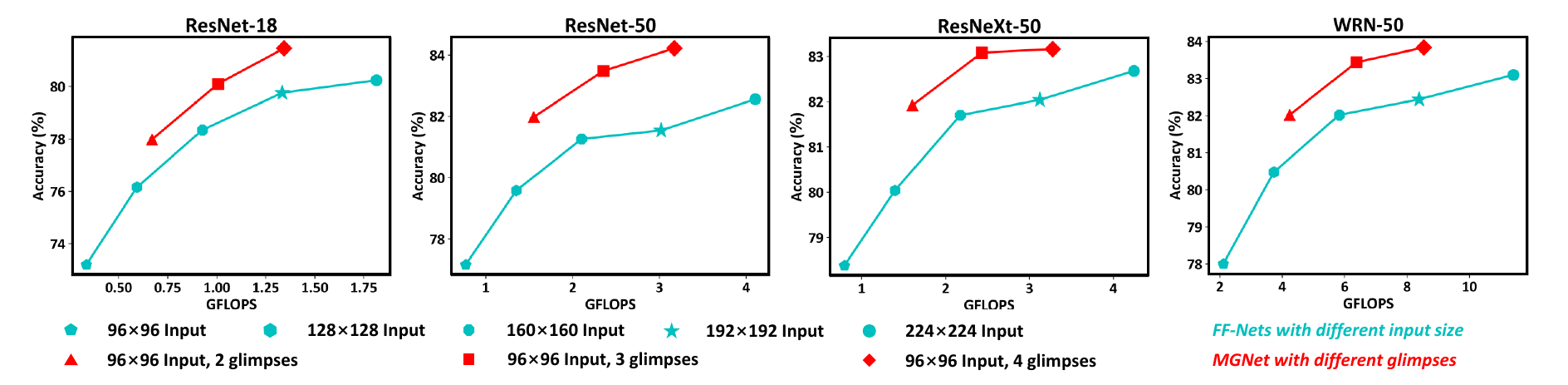}
  \caption{
  Top-1 accuracy (\%) comparison between FF-Nets and MGNet in terms of computational cost on ImageNet100.
MGNet is trained once and exits on a different number of glimpses to show the accuracy of early-exit.
FF-Nets with different input sizes are trained separately to explore the trade-off between the accuracy and computation of the one-pass strategy.
The results show that given the same model capacity, MGNet consistently outperforms FF-Nets among various backbones while having fewer computations.
  }
\label{fig:early-exit}
\end{figure*}

\begin{table}
\vspace{-10pt}
\renewcommand{\arraystretch}{1.4}
\resizebox{\linewidth}{!}{
\begin{tabular}{cc|ccc}
\toprule
\multicolumn{2}{c|}{\textbf{Network}}          & \textbf{GFLOPs} & \textbf{Latency (ms)} & \textbf{Accuracy (\%)} \\
\hline
\hline
\multirow{2}{*}{ResNet-18}  & FF-Net & 1.815     & 87.7          & 80.24       \\
                            & MGNet  & \textbf{1.343}     & \textbf{59.4}          & \textbf{81.46}       \\
\hline
\multirow{2}{*}{ResNet-50}  & FF-Net & 4.104     & 240.1          & 82.56       \\
                            & MGNet  & \textbf{3.172}     & \textbf{167.6}          & \textbf{84.22}       \\
\hline
\multirow{2}{*}{ResNeXt-50} & FF-Net & 4.246     & 313.1          & 82.68       \\
                            & MGNet  & \textbf{3.276}     & \textbf{198.3}          & \textbf{83.16}       \\
\hline
\multirow{2}{*}{WRN-50}     & FF-Net & 11.413     & 486.6          & 83.10       \\
                            & MGNet  & \textbf{8.542}     & \textbf{369.0}          & \textbf{83.84}
                            \\
\bottomrule
\end{tabular}
}
\caption{GFLOPs and inference latency on ImageNet100.}
\label{flops}
\vspace{-10pt}
\end{table}

In this section, we evaluate MGNet on ImageNet100, which is the first 100 classes of ImageNet~\cite{deng2009imagenet}.
We demonstrate some experiments on toy datasets in Appendix \ref{toy_dataset} to better understand how the RDA mechanism works.

We implement FF-Net as a special case of MGNet with the number of glimpses $T = 1$ and downsampling factor $M = 1$, which means the Glimpse Generator performs identity transformation without downsampling.
As our comparison does not depend on backbone architecture, we evaluate it with ResNet-18~\cite{DBLP:conf/cvpr/HeZRS16}, ResNet-50~\cite{DBLP:conf/cvpr/HeZRS16}, ResNeXt-50~\cite{DBLP:conf/cvpr/XieGDTH17}, and WRN-50~\cite{DBLP:conf/bmvc/ZagoruykoK16} backbones.
To ensure the models' convergence to sufficiently demonstrate their capability, we train both FF-Nets and MGNet in 400 epochs with SGD. The peak learning rate is set to be 0.1 using a one-cycle scheduler~\cite{smith2019super}. For data augmentation, we train models with Auto Augmentation~\cite{DBLP:journals/corr/abs-1805-09501}. For MGNet, we set total glimpses $T = 4$ and downsampling factor $M= 7/3$, which still requires less computation than baseline. The hyper-parameter $\alpha$ is set to be 0.6, $s$ is 0.02 for ResNet-18, and 0.01 otherwise.

We present a fair comparison in terms of the parameter numbers, backbone architecture, training settings, and computational cost.
The following experiments show the potential of MGNet to simultaneously reduce computation, improve adversarial robustness, enhance general robustness and be more interpretable in real-world datasets. 
Visualization of success and failure cases are shown in Appendix \ref{visualize}.

\subsubsection{Early-Exit} \label{sec4_2_1}

Early-exit allows a model to be trained once and specialized for efficient deployment, addressing the challenge of efficient inference across resource-constrained devices such as edge-devices ~\cite{DBLP:journals/comsur/WangHLNYC20}.
MGNet is designed to process multi-glimpse sequentially; hence it can naturally early-exit without adding any overhead.

Table~\ref{flops} shows that \textit{given the same model capacity}, MGNet with four 96 $\times$ 96 glimpses always outperforms FF-Nets with standard 224 $\times$ 224 inputs while holding less computation.
For the latency in the practical usage, we are testing on Intel Xeon E5-2650 without GPU.
Additionally, since the input is smaller for each forward pass, MGNet requires noticeably less memory (e.g., reduce by 26.4\% in ResNet-18).
Therefore, the acceleration is more prominent when the memory resources are limited.
We further demonstrate the early exits' accuracy of the same MGNet and train FF-Nets individually with various input sizes to explore the trade-off between these two manners' computational cost and performance.
We observe that RDA mechanisms can consistently outperform the one-pass manner among various backbones.
As shown in Figure~\ref{fig:early-exit}, with the same backbone ResNet-50, MGNet with four 96 $\times$ 96 glimpses outperforms FF-Net with a full 224 $\times$ 224 input by 1.66\% accuracy, while the computation is only about 77.28\% of the latter.
For ResNeXt-50, MGNet with two 96 $\times$ 96 glimpses matches the performance of FF-Net with 192 $\times$ 192 input while requiring only 51.36\% computation.
\textbf{This experiment shows that an image classifier can be more efficient and effective by including RDA mechanisms.}

\subsubsection{Common Corruptions}  \label{sec4_2_3}

The models we train on clean data are directly evaluated on the common corruptions benchmark~\cite{hendrycks2019benchmarking} (reduced to 100 classes) ImageNet100-C, which consists of 15 different corruption types generated algorithmically from noise, blur, weather, and digital categories.
Each corruption type has five severity levels, so the total number of corruption types is 75.

Table~\ref{table2} shows that MGNet yields a substantial improvement in general robustness compared to FF-Nets. For example, MGNet with ResNet-18 backbone with three glimpses increases the average accuracy by 6.56\% compared to FF-Nets, while the computational cost is merely 55\% of the latter. On average, MGNet with two glimpses outperforms FF-Nets by 4.76\% with only 36.9\% computational cost.
\textbf{The progress of MGNet perceiving from a rough overview to detailed parts makes it more robust, even with a single glimpse.}

\begin{table*}[t!]
\centering 
\setlength\tabcolsep{1pt}
\renewcommand{\arraystretch}{1.4}
\resizebox{\linewidth}{!}{
\begin{tabular}{c|cl|cc|ccc|cccc|cccc|cccc}
\toprule  
\multicolumn{5}{c}{} & \multicolumn{3}{c}{\textbf{Noise}} & \multicolumn{4}{c}{\textbf{Blur}} & \multicolumn{4}{c}{\textbf{Weather}} & \multicolumn{4}{c}{\textbf{Digital}}\\
\hline
\cline{1-20}
\multicolumn{3}{c|}{\textbf{Network}} & \textbf{GFLOPs} & \textbf{Average} & Gaussian & Shot & Impulse & Defocus & Glass & Motion & Zoom & Snow    & Frost & Fog & Brightness & Contrast & Elastic & Pixelate & JPEG\\
\hline
\hline
\multirow{4}{*}{ResNet-18} & \multicolumn{2}{c|}{FF-Nets} 
                                       & 1.8146                           & 46.21                             & 36          & 37          & 32          & 28          & 35          & 42          & 41          & 41          & 47          & \textbf{62} & 71          & 52          & 60          & 59          & 52   \\
\cline{2-20}
 & \multirow{3}{*}{MGNet} & 1-glimpse  & 0.3342                           & 50.56                             & 44          & 41          & 39          & \textbf{38} & \textbf{48} & 48          & 43          & 37          & 47          & 50          & 69          & 56          & 66          & 70          & \textbf{63} \\
 &                        & 2-glimpse & 0.6695                           & 52.77                             & \textbf{45} & \textbf{43} & \textbf{40} & \textbf{38} & \textbf{48} & \textbf{49} & 47          & 41          & \textbf{51} & 57          & 73          & \textbf{58} & 68          & \textbf{71} & \textbf{63} \\
 &                        & 3-glimpse & 1.0058                           & \textbf{53.23}                    & \textbf{45} & \textbf{43} & \textbf{40} & \textbf{38} & 47          & \textbf{49} & \textbf{49} & \textbf{43} & \textbf{51} & 59          & \textbf{74} & \textbf{58} & \textbf{69} & \textbf{71} & 62          \\
\hline
\cline{1-20}
\multirow{3}{*}{ResNet-50} & \multicolumn{2}{c|}{FF-Nets} 
                                       & 4.1042                           & 53.24                             & 46          & 46          & 42          & 37          & 43          & 49          & 49          & \textbf{47} & 54          & \textbf{64} & 76          & 60          & 64          & 64          & 59          \\
\cline{2-20}
 & \multirow{2}{*}{MGNet} & 1-glimpse  & 0.7677                           & 55.03                             & 50          & 46          & 46          & 43          & \textbf{53} & 50          & 47          & 42          & 53          & 53          & 73          & 60          & 70          & \textbf{73} & 66          \\
 &                        & 2-glimpse & 1.5523                           & \textbf{57.36}                    & \textbf{51} & \textbf{48} & \textbf{47} & \textbf{43} & \textbf{53} & \textbf{52} & \textbf{54} & \textbf{47} & \textbf{56} & 60          & \textbf{77} & \textbf{61} & \textbf{72} & \textbf{73} & \textbf{67} \\
\hline
\cline{1-20}
\multirow{3}{*}{ResNeXt-50} & \multicolumn{2}{c|}{FF-Nets} 
                                       & 4.2455                           & 53.01                             & 47          & 47          & 42          & 37          & 42          & 47          & 47          & \textbf{48} & 54          & \textbf{63} & \textbf{76} & \textbf{59} & 64          & 62          & 58          \\
\cline{2-20}
 & \multirow{2}{*}{MGNet} & 1-glimpse  & 0.7937                           & 55.57                             & \textbf{51} & 49          & 47          & \textbf{43} & \textbf{52} & 50          & 46          & 45          & 56          & 54          & 74          & \textbf{59} & 70          & \textbf{73} & \textbf{65} \\
 &                        & 2-glimpse & 1.6042                           & \textbf{57.13}                    & \textbf{51} & \textbf{50} & \textbf{49} & 42          & \textbf{52} & \textbf{51} & \textbf{52} & \textbf{48} & \textbf{58} & 59          & \textbf{76} & \textbf{59} & \textbf{71} & \textbf{73} & \textbf{65} \\
\hline
\cline{1-20}
\multirow{3}{*}{WRN-50} & \multicolumn{2}{c|}{FF-Nets}  
                                       & 11.413                           & 54.75                             & 48          & 49          & 45          & 39          & 46          & 49          & 49          & \textbf{49} & 55          & \textbf{64} & \textbf{77} & 61          & 65          & 66          & 60          \\
\cline{2-20}
 & \multirow{2}{*}{MGNet} & 1-glimpse  & 2.1101                           & 56.76                             & 53          & 50          & 49          & 45          & 55          & 51          & 48          & 44          & 55          & 56          & 74          & 60          & 70          & 74          & 67          \\
 &                        & 2-glimpse & 4.2372                           & \textbf{59.02}                    & \textbf{54} & \textbf{51} & \textbf{50} & \textbf{46} & \textbf{55} & \textbf{52} & \textbf{55} & \textbf{49} & \textbf{58} & 62          & \textbf{77} & \textbf{62} & \textbf{73} & \textbf{75} & \textbf{68} \\
\bottomrule
\end{tabular}
}
\caption{Top-1 accuracy (\%) evaluation of MGNet and FF-Nets on ImageNet100-C.}
\label{table2}
\vspace{-10pt}
\end{table*}

\subsubsection{Adversarial Robustness}  \label{sec4_2_2}

Recent work show that deep neural networks can be simply fooled by adversarial examples~\cite{szegedy2013intriguing, goodfellow2014explaining}.
In this section, we compare the adversarial robustness between FF-Nets and MGNet without adversarial training~\cite{madry2017towards}.

FGSM~\cite{goodfellow2014explaining} is one of the most popular methods to generate adversarial examples during a single iteration,

\begin{equation}
\boldsymbol{x} + \epsilon \cdot \mathrm{sgn}\big(\nabla_{\boldsymbol{x}}L(\theta, x, y)\big),
\end{equation}

\noindent where $\boldsymbol{x}$ is an input image, $\boldsymbol{y}$ is the label, $\theta$ denotes the parameters, $L$ is the loss function, $\mathrm{sgn}$ returns the sign, and $\epsilon$ is the attack step size.
PGD~\cite{madry2017towards} is an iterative variant of FGSM,

\begin{equation}
\boldsymbol{x}^{k+1} = \Pi_{\boldsymbol{x}+\mathcal{S}} \  \boldsymbol{x}^k + \epsilon \cdot \mathrm{sgn}\big(\nabla_{\boldsymbol{x}}L(\theta, \boldsymbol{x}, \boldsymbol{y})\big),
\end{equation}

\begin{figure}
\begin{center}
\includegraphics[width=\linewidth]{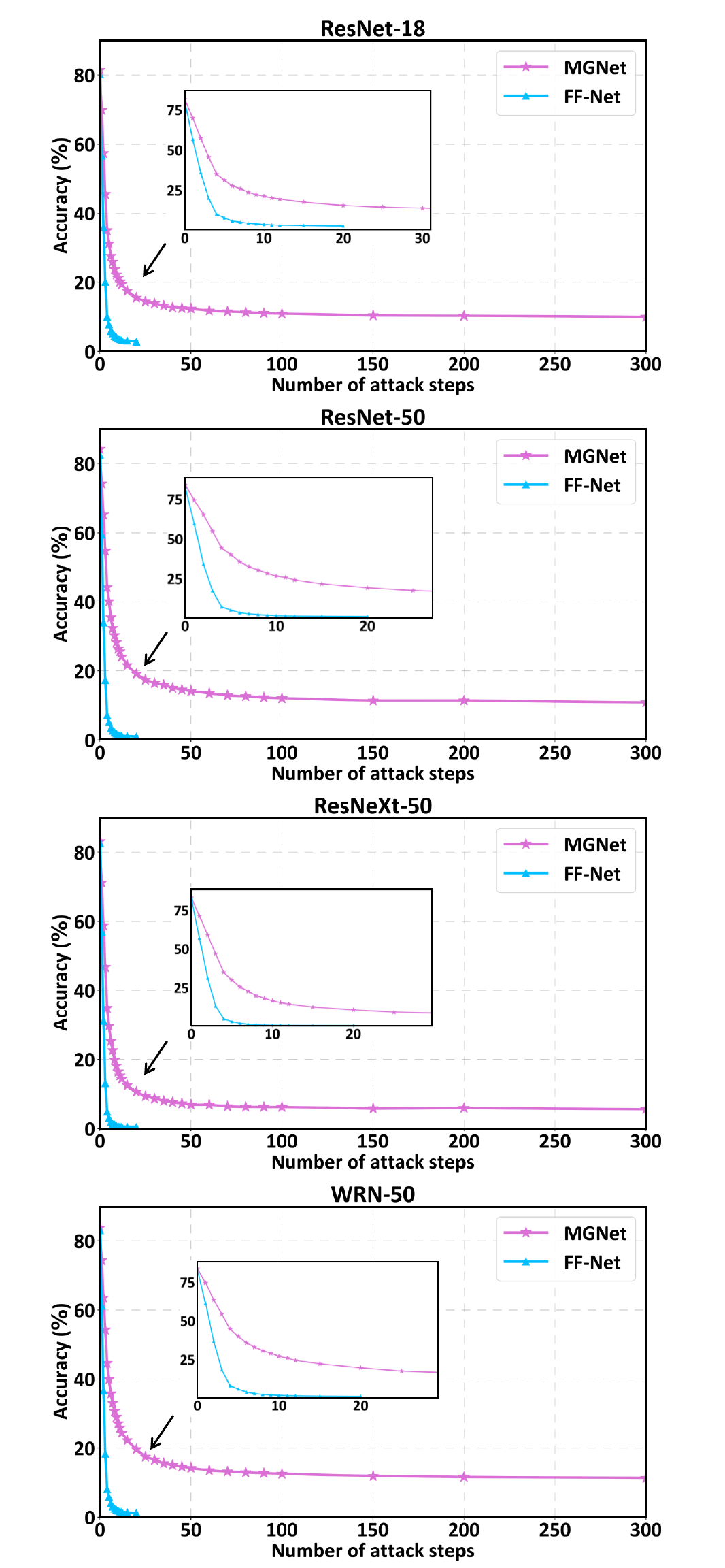}
\end{center}
   \caption{The top-1 accuracy performance comparison over different numbers of PGD attacks without adversarial training on ImageNet100.
   }
\label{fig:attack}
\end{figure}

\noindent where $k$ is the iteration index and $\mathcal{S}$ denotes the set of perturbations that formalizes the manipulative power of the adversary. In the following experiments, we consider the PGD attacks with $4/255$ $\ell_{\infty}$-bounded and step size $\epsilon = 1/255$ on different numbers of steps.

As shown in Figure~\ref{fig:attack}, with the same strength of the PGD attacks, the adversarial robustness of MGNet significantly outperforms FF-Nets. For example, with four attack steps, the top-1 accuracy of FF-Net with ResNet-50 drastically drops to 7.6\%, while MGNet still maintains 44.2\%.
Even with 300 attack steps, the accuracy of MGNet still maintains 10.86\% while FF-Nets drops to 0.96\% with only 20 attack steps.
The result is consistent across various backbones.
We infer that the increment of robustness may come from the ensemble, but MGNet even requires less computational cost than a single pass of FF-Nets.
Note that we intend to show the intrinsic feature of MGNet against adversarial attacks rather than propose a defense method. Besides, the one-stage end-to-end trainable property allows MGNet to be combined with various adversarial defense methods to achieve higher adversarial robustness.

\subsection{Tiny ImageNet}

We evaluate MGNet on Tiny ImageNet~\cite{le2015tiny} to explore the performance on images with lower resolution.
Tiny ImageNet is a subset of ImageNet. It includes 200 distinct categories, and each contains 500 training images, 50 validation images, and 50 test images. All the images are resized to 64 $\times$ 64 pixels, where the original size is 224 $\times$ 224 pixels on ImageNet.

\begin{table}
\begin{threeparttable}
\renewcommand{\arraystretch}{1.2}
\vspace{5pt}

\begin{tabular}{lc|ccc}
\toprule
\multicolumn{2}{c|}{\textbf{Network}}          & \textbf{GFLOPs} & \textbf{Accuracy (\%)} \\
\hline
\hline
\multirow{2}{*}{ResNet-18}  & Vanilla~\cite{sun2016resnet}   & 0.1497     & 52.40       \\
                            & Ours  & 0.1497     & 53.97       \\
\multirow{2}{*}{ResNet-18}\tnote{\dag}  & FF-Net & 0.5657     & 57.14       \\
                            & MGNet  & 0.4301     & 57.72       \\
\hline
\multirow{2}{*}{ResNet-34} & Vanilla~\cite{sun2016resnet}  & 0.3009     & 53.20       \\
                            & Ours  & 0.3009     & 55.08       \\
\multirow{2}{*}{ResNet-34}\tnote{\dag}     & FF-Net & 1.1705    & 58.71       \\
                            & MGNet  & 0.8837     & 58.38
                            \\
\bottomrule
\end{tabular}

\begin{tablenotes}
\footnotesize
\item[\dag] No max-pooling layer followed by the first convolutional layer.
\end{tablenotes}
\caption{GFLOPs and accuracy (\%) evaluation on Tiny ImageNet.
}
\label{tiny}
\vspace{10pt}

\end{threeparttable}
\vspace{-20pt}
\end{table}

We select downsampling factor $M = 2$ and total glimpses $T = 3$ for MGNet to make an appropriate comparison with FF-Nets.
In this setting, MGNet will receive three 32 $\times$ 32 pixels glimpses while FF-Nets, as usual, will receive a 64 $\times$ 64 pixels image.
We first compare our baseline implementation with~\cite{sun2016resnet}.
Next, same as~\cite{wu2017tiny}, we remove the max-pooling layer followed by the first convolutional layer as we will reduce the input image size further to 32 $\times$ 32 pixels.
Note that these networks are initially designed for 224 $\times$ 224 pixels images.
We use the notation $\dagger$ to mark modified networks that we select as the backbones to compare FF-Nets and MGNet.

As shown in Table~\ref{tiny}, the feedforward baselines of our implementation are slightly higher than~\cite{sun2016resnet} baselines.
It can benefit from our learning-rate scheduler choice and the larger training epochs that ensure the models are fully converged.
In these experiments, we show the potential of RDA mechanism to reduce computation while maintaining accuracy in smaller image scales.
For example, using ResNet-18$^\dagger$ as the backbone, FF-Net and MGNet achieve a comparable accuracy while the latter requires only 76\% FLOPs. 
\noindent This improvement may not be so significant at larger image scales.
Nevertheless, we claim these results are reasonable because the smaller the image is, the less redundant computing is spent on unimportant regions.

\vspace{-2.5pt}
\section{Conclusion}

In this paper, we explore the capability of a recurrent downsampled attention mechanism based model for image classification.
MGNet achieves comparable predictive performance on ImageNet100 while holding several benefits: 1) requires less computation amount; 2) can early-exit on-the-fly; 3) is intrinsically more robust against adversarial attacks and common corruptions; and 4) explicitly informs more spatial information.
Furthermore, we can directly train MGNet in an end-to-end manner from scratch.

Although we intuitively propose to train MGNet by gradient re-scaling, it harms the convergence speed, and such that we cannot afford to explore MGNet on ImageNet dataset.
Future work can focus on tackling this problem or improving MGNet submodules.

Beyond that, there is no apparent limitation for MGNet to be combined with recent work such as pruning, quantization, knowledge distillation, and adversarial defense methods to achieve more promising performance.
We hope that this work will spur the related research direction that focuses on the exploration of recurrent downsampled attention mechanism to improve vision models further.

{\small
\bibliographystyle{ieee_fullname}
\bibliography{main}

\begin{thebibliography}{10}\itemsep=-1pt

\bibitem{DBLP:journals/corr/BahdanauCB14}
Dzmitry Bahdanau, Kyunghyun Cho, and Yoshua Bengio.
\newblock Neural machine translation by jointly learning to align and
  translate.
\newblock In {\em Int. Conf. Learn. Represent. (ICLR)}, 2015.

\bibitem{NMT}
Dzmitry Bahdanau, Kyunghyun Cho, and Yoshua Bengio.
\newblock Neural machine translation by jointly learning to align and
  translate.
\newblock In {\em Int. Conf. Learn. Represent. (ICLR)}, 2015.

\bibitem{chen2017deeplab}
Liang{-}Chieh Chen, George Papandreou, Iasonas Kokkinos, Kevin Murphy, and
  Alan~L. Yuille.
\newblock Deeplab: Semantic image segmentation with deep convolutional nets,
  atrous convolution, and fully connected crfs.
\newblock {\em IEEE Trans. Pattern Anal. Mach. Intell. (TPAMI)},
  40(4):834--848, 2018.

\bibitem{DBLP:conf/iclr/ChoiEL17}
Yoojin Choi, Mostafa El{-}Khamy, and Jungwon Lee.
\newblock Towards the limit of network quantization.
\newblock In {\em Int. Conf. Learn. Represent. (ICLR)}, 2017.

\bibitem{DBLP:journals/corr/abs-1805-09501}
Ekin~Dogus Cubuk, Barret Zoph, Dandelion Man{\'{e}}, Vijay Vasudevan, and
  Quoc~V. Le.
\newblock Autoaugment: Learning augmentation policies from data.
\newblock {\em CoRR}, abs/1805.09501, 2018.

\bibitem{deng2009imagenet}
Jia Deng, Wei Dong, Richard Socher, Li{-}Jia Li, Kai Li, and Fei{-}Fei Li.
\newblock Imagenet: {A} large-scale hierarchical image database.
\newblock In {\em IEEE Conf. Comput. Vis. Pattern Recog. (CVPR)}, pages
  248--255, 2009.

\bibitem{DBLP:conf/acl/DevlinCFGDHZM15}
Jacob Devlin, Hao Cheng, Hao Fang, Saurabh Gupta, Li Deng, Xiaodong He,
  Geoffrey Zweig, and Margaret Mitchell.
\newblock Language models for image captioning: The quirks and what works.
\newblock pages 100--105, 2015.

\bibitem{DBLP:conf/nips/EslamiHWTSKH16}
S.~M.~Ali Eslami, Nicolas Heess, Theophane Weber, Yuval Tassa, David
  Szepesvari, Koray Kavukcuoglu, and Geoffrey~E. Hinton.
\newblock Attend, infer, repeat: Fast scene understanding with generative
  models.
\newblock In {\em Adv. Neural Inform. Process. Syst. (NIPS)}, pages 3225--3233,
  2016.

\bibitem{Zheng_2017_ICCV}
Jianlong Fu, Heliang Zheng, and Tao Mei.
\newblock Look closer to see better: Recurrent attention convolutional neural
  network for fine-grained image recognition.
\newblock In {\em IEEE Conf. Comput. Vis. Pattern Recog. (CVPR)}, pages
  4476--4484, 2017.

\bibitem{DBLP:conf/icml/GehringAGYD17}
Jonas Gehring, Michael Auli, David Grangier, Denis Yarats, and Yann~N. Dauphin.
\newblock Convolutional sequence to sequence learning.
\newblock In {\em Proc. Int. Conf. Mach. Learn. (ICML)}, volume~70, pages
  1243--1252, 2017.

\bibitem{geirhos2018imagenet}
Robert Geirhos, Patricia Rubisch, Claudio Michaelis, Matthias Bethge, Felix~A.
  Wichmann, and Wieland Brendel.
\newblock Imagenet-trained cnns are biased towards texture; increasing shape
  bias improves accuracy and robustness.
\newblock In {\em Int. Conf. Learn. Represent. (ICLR)}, 2019.

\bibitem{girshick2014rich}
Ross~B. Girshick, Jeff Donahue, Trevor Darrell, and Jitendra Malik.
\newblock Rich feature hierarchies for accurate object detection and semantic
  segmentation.
\newblock In {\em IEEE Conf. Comput. Vis. Pattern Recog. (CVPR)}, pages
  580--587, 2014.

\bibitem{goodfellow2014explaining}
Ian~J. Goodfellow, Jonathon Shlens, and Christian Szegedy.
\newblock Explaining and harnessing adversarial examples.
\newblock In {\em Int. Conf. Learn. Represent. (ICLR)}, 2015.

\bibitem{gregor2015draw}
Karol Gregor, Ivo Danihelka, Alex Graves, Danilo~Jimenez Rezende, and Daan
  Wierstra.
\newblock {DRAW:} {A} recurrent neural network for image generation.
\newblock In {\em Proc. Int. Conf. Mach. Learn. (ICML)}, volume~37, pages
  1462--1471, 2015.

\bibitem{han2015deep}
Song Han, Huizi Mao, and William~J Dally.
\newblock Deep compression: Compressing deep neural networks with pruning,
  trained quantization and huffman coding.
\newblock {\em arXiv preprint arXiv:1510.00149}, 2015.

\bibitem{DBLP:conf/cvpr/HaqueAF16}
Albert Haque, Alexandre Alahi, and Li Fei{-}Fei.
\newblock Recurrent attention models for depth-based person identification.
\newblock In {\em IEEE Conf. Comput. Vis. Pattern Recog. (CVPR)}, pages
  1229--1238, 2016.

\bibitem{DBLP:conf/iccv/HeZRS15}
Kaiming He, Xiangyu Zhang, Shaoqing Ren, and Jian Sun.
\newblock Delving deep into rectifiers: Surpassing human-level performance on
  imagenet classification.
\newblock In {\em Int. Conf. Comput. Vis. (ICCV)}, pages 1026--1034, 2015.

\bibitem{he2016deep}
Kaiming He, Xiangyu Zhang, Shaoqing Ren, and Jian Sun.
\newblock Deep residual learning for image recognition.
\newblock In {\em IEEE Conf. Comput. Vis. Pattern Recog. (CVPR)}, pages
  770--778, 2016.

\bibitem{DBLP:conf/cvpr/HeZRS16}
Kaiming He, Xiangyu Zhang, Shaoqing Ren, and Jian Sun.
\newblock Deep residual learning for image recognition.
\newblock In {\em IEEE Conf. Comput. Vis. Pattern Recog. (CVPR)}, pages
  770--778, 2016.

\bibitem{hendrycks2020many}
Dan Hendrycks, Steven Basart, Norman Mu, Saurav Kadavath, Frank Wang, Evan
  Dorundo, Rahul Desai, Tyler Zhu, Samyak Parajuli, Mike Guo, Dawn Song, Jacob
  Steinhardt, and Justin Gilmer.
\newblock The many faces of robustness: {A} critical analysis of
  out-of-distribution generalization.
\newblock {\em CoRR}, abs/2006.16241, 2020.

\bibitem{hendrycks2019benchmarking}
Dan Hendrycks and Thomas~G. Dietterich.
\newblock Benchmarking neural network robustness to common corruptions and
  perturbations.
\newblock In {\em Int. Conf. Learn. Represent. (ICLR)}, 2019.

\bibitem{hendrycks2019augmix}
Dan Hendrycks, Norman Mu, Ekin~Dogus Cubuk, Barret Zoph, Justin Gilmer, and
  Balaji Lakshminarayanan.
\newblock Augmix: {A} simple data processing method to improve robustness and
  uncertainty.
\newblock In {\em Int. Conf. Learn. Represent. (ICLR)}, 2020.

\bibitem{hinton2015distilling}
Geoffrey~E. Hinton, Oriol Vinyals, and Jeffrey Dean.
\newblock Distilling the knowledge in a neural network.
\newblock {\em CoRR}, abs/1503.02531, 2015.

\bibitem{DBLP:journals/corr/abs-1810-04020}
Md.~Zakir Hossain, Ferdous Sohel, Mohd~Fairuz Shiratuddin, and Hamid Laga.
\newblock A comprehensive survey of deep learning for image captioning.
\newblock {\em CoRR}, abs/1810.04020, 2018.

\bibitem{DBLP:journals/corr/HowardZCKWWAA17}
Andrew~G. Howard, Menglong Zhu, Bo Chen, Dmitry Kalenichenko, Weijun Wang,
  Tobias Weyand, Marco Andreetto, and Hartwig Adam.
\newblock Mobilenets: Efficient convolutional neural networks for mobile vision
  applications.
\newblock {\em CoRR}, abs/1704.04861, 2017.

\bibitem{DBLP:journals/pami/HuSASW20}
Jie Hu, Li Shen, Samuel Albanie, Gang Sun, and Enhua Wu.
\newblock Squeeze-and-excitation networks.
\newblock {\em IEEE Trans. Pattern Anal. Mach. Intell. (TPAMI)},
  42(8):2011--2023, 2020.

\bibitem{huang2017densely}
Gao Huang, Zhuang Liu, and Kilian~Q. Weinberger.
\newblock Densely connected convolutional networks.
\newblock {\em CoRR}, abs/1608.06993, 2016.

\bibitem{courbariaux2016binarized}
Itay Hubara, Matthieu Courbariaux, Daniel Soudry, Ran El{-}Yaniv, and Yoshua
  Bengio.
\newblock Binarized neural networks.
\newblock In {\em Adv. Neural Inform. Process. Syst. (NIPS)}, pages 4107--4115,
  2016.

\bibitem{jaderberg2015spatial}
Max Jaderberg, Karen Simonyan, Andrew Zisserman, and Koray Kavukcuoglu.
\newblock Spatial transformer networks.
\newblock In {\em Adv. Neural Inform. Process. Syst. (NIPS)}, pages 2017--2025,
  2015.

\bibitem{DBLP:journals/corr/abs-2103-03206}
Andrew Jaegle, Felix Gimeno, Andrew Brock, Andrew Zisserman, Oriol Vinyals, and
  Jo{\~{a}}o Carreira.
\newblock Perceiver: General perception with iterative attention.
\newblock {\em CoRR}, abs/2103.03206, 2021.

\bibitem{krizhevsky2017imagenet}
Alex Krizhevsky, Ilya Sutskever, and Geoffrey~E. Hinton.
\newblock Imagenet classification with deep convolutional neural networks.
\newblock {\em Commun. {ACM}}, 60(6):84--90, 2017.

\bibitem{DBLP:journals/corr/abs-1912-10773}
Sampo Kuutti, Richard Bowden, Yaochu Jin, Phil Barber, and Saber Fallah.
\newblock A survey of deep learning applications to autonomous vehicle control.
\newblock {\em CoRR}, abs/1912.10773, 2019.

\bibitem{Glimpse}
Hugo Larochelle and Geoffrey~E. Hinton.
\newblock Learning to combine foveal glimpses with a third-order boltzmann
  machine.
\newblock In {\em Adv. Neural Inform. Process. Syst. (NIPS)}, pages 1243--1251,
  2010.

\bibitem{le2015tiny}
Ya Le and Xuan Yang.
\newblock Tiny imagenet visual recognition challenge.
\newblock {\em CS 231N}, 7(7):3, 2015.

\bibitem{lecun-mnisthandwrittendigit-2010}
Yann LeCun and Corinna Cortes.
\newblock {MNIST} handwritten digit database.
\newblock 2010.

\bibitem{DBLP:conf/nips/LiaoLSWHDUZ19}
Renjie Liao, Yujia Li, Yang Song, Shenlong Wang, William~L. Hamilton, David
  Duvenaud, Raquel Urtasun, and Richard~S. Zemel.
\newblock Efficient graph generation with graph recurrent attention networks.
\newblock In {\em Adv. Neural Inform. Process. Syst. (NIPS)}, pages 4257--4267,
  2019.

\bibitem{Saccadic}
S.~P. Liversedge and J. Findlay.
\newblock Saccadic eye movements and cognition.
\newblock {\em Trends in Cognitive Sciences}, 4:6--14, 2000.

\bibitem{long2015fully}
Jonathan Long, Evan Shelhamer, and Trevor Darrell.
\newblock Fully convolutional networks for semantic segmentation.
\newblock In {\em IEEE Conf. Comput. Vis. Pattern Recog. (CVPR)}, pages
  3431--3440, 2015.

\bibitem{DBLP:conf/aaai/LuoZLWT16}
Ping Luo, Zhenyao Zhu, Ziwei Liu, Xiaogang Wang, and Xiaoou Tang.
\newblock Face model compression by distilling knowledge from neurons.
\newblock In {\em AAAI Conf. Artif. Intell. (AAAI)}, pages 3560--3566, 2016.

\bibitem{DBLP:conf/eccv/MaZZS18}
Ningning Ma, Xiangyu Zhang, Hai{-}Tao Zheng, and Jian Sun.
\newblock Shufflenet {V2:} practical guidelines for efficient {CNN}
  architecture design.
\newblock In {\em Eur. Conf. Comput. Vis. (ECCV)}, volume 11218, pages
  122--138, 2018.

\bibitem{madry2017towards}
Aleksander Madry, Aleksandar Makelov, Ludwig Schmidt, Dimitris Tsipras, and
  Adrian Vladu.
\newblock Towards deep learning models resistant to adversarial attacks.
\newblock In {\em Int. Conf. Learn. Represent. (ICLR)}, 2018.

\bibitem{RAM}
Volodymyr Mnih, Nicolas Heess, Alex Graves, and Koray Kavukcuoglu.
\newblock Recurrent models of visual attention.
\newblock {\em CoRR}, abs/1406.6247, 2014.

\bibitem{DBLP:conf/iccv/Mustafa0HGS019}
Aamir Mustafa, Salman~H. Khan, Munawar Hayat, Roland Goecke, Jianbing Shen, and
  Ling Shao.
\newblock Adversarial defense by restricting the hidden space of deep neural
  networks.
\newblock In {\em Int. Conf. Comput. Vis. (ICCV)}, pages 3384--3393, 2019.

\bibitem{nguyen2015deep}
Anh~Mai Nguyen, Jason Yosinski, and Jeff Clune.
\newblock Deep neural networks are easily fooled: High confidence predictions
  for unrecognizable images.
\newblock In {\em IEEE Conf. Comput. Vis. Pattern Recog. (CVPR)}, pages
  427--436, 2015.

\bibitem{niv2015reinforcement}
Yael Niv, Reka Daniel, Andra Geana, Samuel~J Gershman, Yuan~Chang Leong, Angela
  Radulescu, and Robert~C Wilson.
\newblock Reinforcement learning in multidimensional environments relies on
  attention mechanisms.
\newblock {\em Journal of Neuroscience}, 35(21):8145--8157, 2015.

\bibitem{vision20}
Bruno~A Olshausen.
\newblock 20 years of learning about vision: Questions answered, questions
  unanswered, and questions not yet asked.
\newblock In {\em 20 Years of Computational Neuroscience}, pages 243--270.
  2013.

\bibitem{DBLP:journals/corr/abs-2002-07405}
Yao Qin, Nicholas Frosst, Colin Raffel, Garrison~W. Cottrell, and Geoffrey~E.
  Hinton.
\newblock Deflecting adversarial attacks.
\newblock {\em CoRR}, abs/2002.07405, 2020.

\bibitem{DBLP:conf/eccv/RastegariORF16}
Mohammad Rastegari, Vicente Ordonez, Joseph Redmon, and Ali Farhadi.
\newblock Xnor-net: Imagenet classification using binary convolutional neural
  networks.
\newblock In {\em Eur. Conf. Comput. Vis. (ECCV)}, volume 9908 of {\em Lecture
  Notes in Computer Science}, pages 525--542, 2016.

\bibitem{redmon2016you}
Joseph Redmon, Santosh~Kumar Divvala, Ross~B. Girshick, and Ali Farhadi.
\newblock You only look once: Unified, real-time object detection.
\newblock In {\em IEEE Conf. Comput. Vis. Pattern Recog. (CVPR)}, pages
  779--788, 2016.

\bibitem{ren2015faster}
Shaoqing Ren, Kaiming He, Ross~B. Girshick, and Jian Sun.
\newblock Faster {R-CNN:} towards real-time object detection with region
  proposal networks.
\newblock In {\em Adv. Neural Inform. Process. Syst. (NIPS)}, pages 91--99,
  2015.

\bibitem{Dynamic}
Ronald~A. Rensink.
\newblock The dynamic representation of scenes.
\newblock {\em Visual Cognition}, 7:17 -- 42, 2000.

\bibitem{DBLP:journals/corr/RomeroBKCGB14}
Adriana Romero, Nicolas Ballas, Samira~Ebrahimi Kahou, Antoine Chassang, Carlo
  Gatta, and Yoshua Bengio.
\newblock Fitnets: Hints for thin deep nets.
\newblock In {\em Int. Conf. Learn. Represent. (ICLR)}, 2015.

\bibitem{DBLP:conf/cvpr/SandlerHZZC18}
Mark Sandler, Andrew~G. Howard, Menglong Zhu, Andrey Zhmoginov, and
  Liang{-}Chieh Chen.
\newblock Mobilenetv2: Inverted residuals and linear bottlenecks.
\newblock In {\em IEEE Conf. Comput. Vis. Pattern Recog. (CVPR)}, pages
  4510--4520, 2018.

\bibitem{smith2019super}
Leslie~N Smith and Nicholay Topin.
\newblock Super-convergence: Very fast training of neural networks using large
  learning rates.
\newblock In {\em Artificial Intelligence and Machine Learning for Multi-Domain
  Operations Applications}, volume 11006, page 1100612. International Society
  for Optics and Photonics, 2019.

\bibitem{DBLP:conf/bmvc/SrinivasB15}
Suraj Srinivas and R.~Venkatesh Babu.
\newblock Data-free parameter pruning for deep neural networks.
\newblock In {\em Brit. Mach. Vis. Conf. (BMVC)}, pages 31.1--31.12, 2015.

\bibitem{sun2016resnet}
Lei Sun.
\newblock Resnet on tiny imagenet.
\newblock {\em Submitted on}, 14, 2016.

\bibitem{szegedy2013intriguing}
Christian Szegedy, Wojciech Zaremba, Ilya Sutskever, Joan Bruna, Dumitru Erhan,
  Ian~J. Goodfellow, and Rob Fergus.
\newblock Intriguing properties of neural networks.
\newblock In {\em Int. Conf. Learn. Represent. (ICLR)}, 2014.

\bibitem{DBLP:conf/iclr/UllrichMW17}
Karen Ullrich, Edward Meeds, and Max Welling.
\newblock Soft weight-sharing for neural network compression.
\newblock In {\em Int. Conf. Learn. Represent. (ICLR)}, 2017.

\bibitem{Primate}
David~C Van~Essen and Charles~H Anderson.
\newblock Information processing strategies and pathways in the primate visual
  system.
\newblock {\em An introduction to neural and electronic networks}, 2:45--76,
  1995.

\bibitem{vaswani2017attention}
Ashish Vaswani, Noam Shazeer, Niki Parmar, Jakob Uszkoreit, Llion Jones,
  Aidan~N. Gomez, Lukasz Kaiser, and Illia Polosukhin.
\newblock Attention is all you need.
\newblock In {\em Adv. Neural Inform. Process. Syst. (NIPS)}, pages 5998--6008,
  2017.

\bibitem{DBLP:journals/corr/abs-1805-09019}
Qingzhong Wang and Antoni~B. Chan.
\newblock {CNN+CNN:} convolutional decoders for image captioning.
\newblock {\em CoRR}, abs/1805.09019, 2018.

\bibitem{DBLP:journals/comsur/WangHLNYC20}
Xiaofei Wang, Yiwen Han, Victor C.~M. Leung, Dusit Niyato, Xueqiang Yan, and Xu
  Chen.
\newblock Convergence of edge computing and deep learning: {A} comprehensive
  survey.
\newblock {\em {IEEE} Commun. Surv. Tutorials}, 22(2):869--904, 2020.

\bibitem{wu2017tiny}
Jiayu Wu, Qixiang Zhang, and Guoxi Xu.
\newblock Tiny imagenet challenge.
\newblock {\em Technical Report}, 2017.

\bibitem{DBLP:conf/cvpr/XieGDTH17}
Saining Xie, Ross~B. Girshick, Piotr Doll{\'{a}}r, Zhuowen Tu, and Kaiming He.
\newblock Aggregated residual transformations for deep neural networks.
\newblock In {\em IEEE Conf. Comput. Vis. Pattern Recog. (CVPR)}, pages
  5987--5995, 2017.

\bibitem{DBLP:journals/ijautcomp/XuMLDLTJ20}
Han Xu, Yao Ma, Haochen Liu, Debayan Deb, Hui Liu, Jiliang Tang, and Anil~K.
  Jain.
\newblock Adversarial attacks and defenses in images, graphs and text: {A}
  review.
\newblock {\em Int. J. Autom. Comput.}, 17(2):151--178, 2020.

\bibitem{DBLP:conf/bmvc/ZagoruykoK16}
Sergey Zagoruyko and Nikos Komodakis.
\newblock Wide residual networks.
\newblock In {\em Brit. Mach. Vis. Conf. (BMVC)}, 2016.

\bibitem{DBLP:journals/corr/abs-1805-08318}
Han Zhang, Ian~J. Goodfellow, Dimitris~N. Metaxas, and Augustus Odena.
\newblock Self-attention generative adversarial networks.
\newblock {\em CoRR}, abs/1805.08318, 2018.

\bibitem{zoran2019robust}
Daniel Zoran, Mike Chrzanowski, Po{-}Sen Huang, Sven Gowal, Alex Mott, and
  Pushmeet Kohli.
\newblock Towards robust image classification using sequential attention
  models.
\newblock In {\em IEEE Conf. Comput. Vis. Pattern Recog. (CVPR)}, pages
  9480--9489, 2020.

\end{thebibliography}
}

\newpage
\clearpage
\newpage
\appendix
\section*{Appendix}

\section{Study on Hyper-parameters} \label{hparam}

\begin{figure}[t!]
\begin{center}
\includegraphics[width=\linewidth]{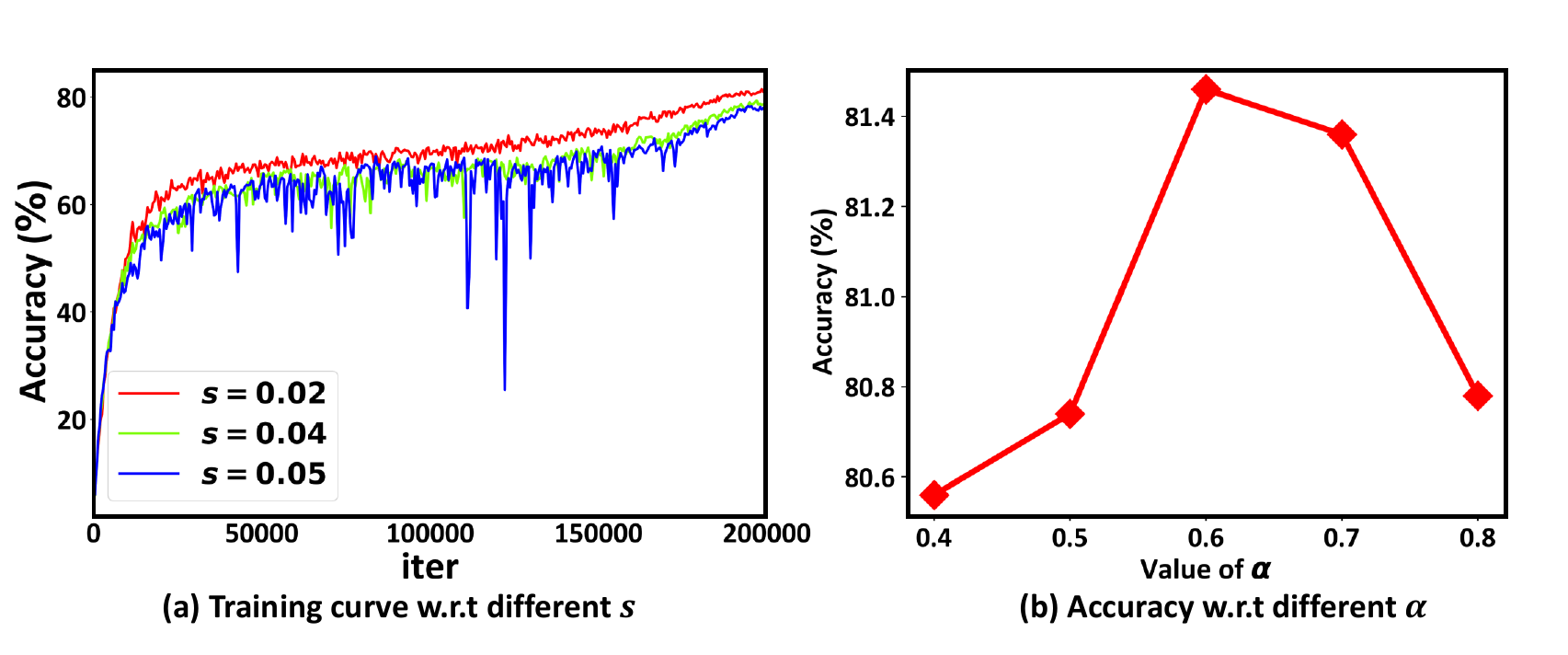}
\end{center}
  \caption{
  Study of different hyper-parameters with ResNet-18 on ImageNet100.
  }
\label{fig:hparam}
\end{figure}

We introduce two hyper-parameters in this work: gradient scaling factor $s$ to tackle the exploding gradient problem in the localization network and $\alpha$ to balance the losses between the two classifiers.
Figure~\ref{fig:hparam} (a) shows that a suitable $s$ works to improve training stability while (b) demonstrates that an appropriate balance between the global classifier and the glimpse classifier boosts accuracy.

\section{Experiments on Toy Datasets} \label{toy_dataset}

We consider the same vanilla CNN backbone for FF-Nets and MGNet in the following experiments, which has four convolutional layers with 16, 32, 64, and 128 filters of kernel size 5, 5, 3, 3, respectively, followed by a global average pooling and a fully-connected layer.

{\bf Translated MNIST (T-MNIST).}
First, we consider the T-MNIST dataset to show the capability of MGNet to capture accurate task-relevant regions.
The data are generated on-the-fly by placing each 28 $\times$ 28 pixels MNIST digit in a random location of a 112 $\times$ 112 blank patch.
We let downsampling factor $M=4$ for our MGNet, which means each glimpse will be 28 $\times$ 28 pixels.
We also train an FF-Net as a baseline, which processes the full-resolution image at once.

\begin{figure}[t!]
\begin{center}
\includegraphics[width=\linewidth]{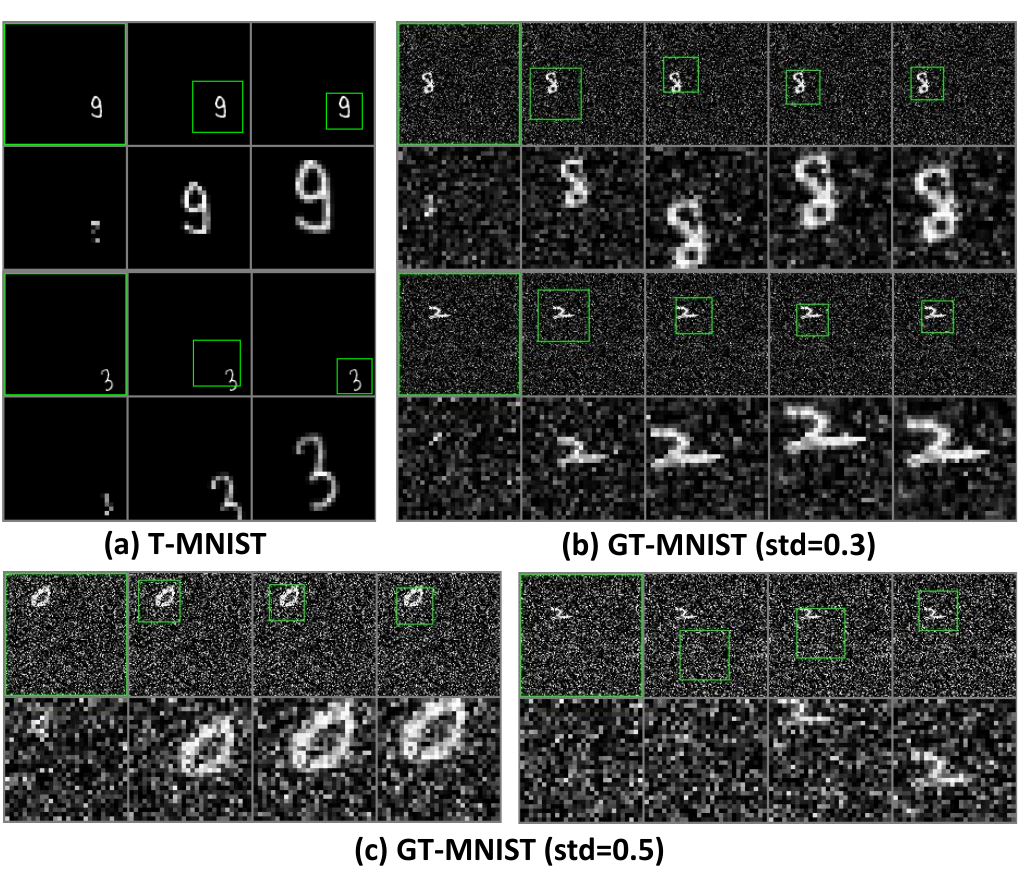}
\end{center}
   \caption{Visualization of the glimpses series generated by MGNet on various datasets.
For each series, the first row shows the original image, each with a green box representing the glimpse-region, while the second row shows the generated glimpses (upsampled for better visualization) sampling from the glimpse-region.
   }
\label{fig:mnist_samples}
\end{figure}

Table~\ref{table1} shows the comparison of the accuracy and computational cost of FF-Net and MGNet.
MGNet stops at number of glimpses $T=3$ as it matches the baseline accuracy, and since it computes on low-dimensionality glimpses, the computation amount can be reduced from 67.95 MFLOPs to 10.619 MFLOPs, which brings $\sim7\times$ computational efficiency boost.
Some samples of generated glimpses are shown in Figure~\ref{fig:mnist_samples} (a) to visualize the glimpse-regions.
Note that we explicitly add an $\ell_2$-norm to the size of the glimpse-region (i.e. ${\left\|a^s-a^s_{\min}\right\|}_2$) to enhance the ability of MGNet to capture precise location without supervised spatial guidance.
\begin{table*}[t!]
\centering
\renewcommand{\arraystretch}{1.2}
\begin{tabular}{c|cl|cc}
\toprule  
\textbf{Dataset} & \multicolumn{2}{c|}{\textbf{Structure}} & \textbf{MFLOPs} & \textbf{Accuracy(\%)} \\
\hline
\hline
\multirow{4}{*}{Translated MNIST } & \multicolumn{2}{c|}{FF-Nets} & 67.950 & 99.48    \\ 
\cline{2-5}
& \multirow{3}{*}{MGNet} & 1-glimpse  & 3.471  & 84.37    \\
&                           & 2-glimpse & 7.011  & 99.31    \\
&                           & 3-glimpse & 10.619 & 99.54    \\
\hline
\cline{1-5}
\multirow{6}{*}{Gaussian Noise T-MNIST} & \multicolumn{2}{c|}{FF-Nets} & 67.950 & 99.10    \\
\cline{2-5}
& \multirow{5}{*}{MGNet} & 1-glimpse     & 3.471  & 63.19    \\
&                           & 2-glimpse & 7.011  & 97.88    \\
&                           & 3-glimpse & 10.619 & 98.87    \\
&                           & 4-glimpse & 14.296 & 99.02    \\
&                           & 5-glimpse & 18.042 &  99.07    \\

\bottomrule
\end{tabular}
\caption{The comparison of the top-1 accuracy (\%) and MFLOPs between FF-Net and MGNet.
FF-Nets sweep the full 112 $\times$ 112 pixels input at once while our MGNet processes several glimpses of 28 $\times$ 28 pixels.
}
\label{table1}
\end{table*}

{\bf Gaussian Noise T-MNIST (GT-MNIST).}
Second, we consider the GT-MNIST dataset, which is generated by adding zero-mean Gaussian noise with a standard deviation of 0.3 to T-MNIST, followed by clipping to maintain the proper image data range. The experiment setup is the same as the previous one. 
As shown in Figure~\ref{fig:mnist_samples} (b),
high-intensity downsampling hinders the capture of a task-relevant region because the first glimpse's spatial information may be ambiguous.
However, by sequentially sampling more glimpses, MGNet matches the baseline with $\sim4\times$ computational efficiency boost as shown in Table~\ref{table1}.
This experiment further explores the ability of MGNet to integrate the information over multiple glimpses.

We further increase the noise intensity to study the behavior of MGNet.
Interestingly, as shown in Figure~\ref{fig:mnist_samples} (c), if the first glimpse does not provide accurate spatial information, the glimpse-region adaptively grows larger to be more perceptive (the second example), and a search is performed to find the task-relevant region.

\begin{figure*}[h!]
\begin{center}
\includegraphics[width=1.0\linewidth]{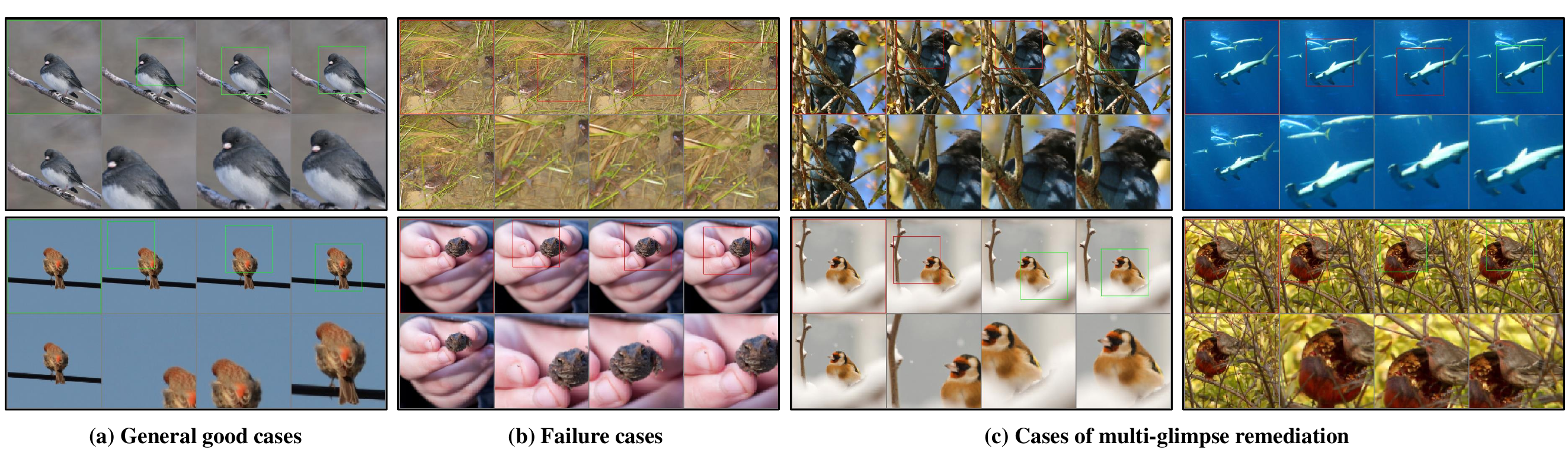}
\end{center}
   \caption{A visualization of the glimpses series generated by MGNet on ImageNet100 validation set.
   A green box denotes that the model has a correct prediction at this glimpse and the red vice versa.
}
\label{fig:imagenet_samples}
\end{figure*}

\section{Visualization of glimpses} \label{visualize}

Since the glimpse-region is dynamically obtained to capture the task-relevant region fields well, the real-world scene's visualization will illustrate this recurrent attention procedure.
We visualize the predicted glimpse-region results taken from ImageNet100 validation dataset in Figure~\ref{fig:imagenet_samples}, which includes three common cases:
1) Figure~\ref{fig:imagenet_samples} (a) shows the examples that the model has correct prediction from the first glimpse while still seeking a precise glimpse-region.
This is because the Glimpse Classifier guides every glimpse, making the model more interpretable that explicitly provides more meaningful task-relevant regions;
2) Figure~\ref{fig:imagenet_samples} (b) shows the general failure cases.
We find that MGNet sometimes fails due to the over-complicated and confusing scene or the tiny object size;
3) Figure~\ref{fig:imagenet_samples} (c) shows the progress that MGNet first predicts wrong but later corrects itself.
Note that in some cases, a more precise glimpse-region is found later, showing the model integrates the information well during the iteration. In other cases, the model looks at a similar region but changes the prediction. We infer this improvement comes from the inherent ensemble feature.

\end{document}